\documentclass[sigconf]{acmart}

\usepackage{algorithm}
\usepackage{framed}
\usepackage{xcolor}
\usepackage{algorithmic}
\usepackage{enumitem}
\usepackage{multirow}
\usepackage{tabularx}

\AtBeginDocument{%
  }

\copyrightyear{2024}
\acmYear{2024}
\setcopyright{rightsretained}
\acmConference[ICAIF '24]{5th ACM International Conference on AI in Finance}{November 14--17, 2024}{Brooklyn, NY, USA}
\acmBooktitle{5th ACM International Conference on AI in Finance (ICAIF '24), November 14--17, 2024, Brooklyn, NY, USA}
\acmDOI{10.1145/3677052.3698645}
\acmISBN{979-8-4007-1081-0/24/11}

\begin{document}

\title{Enhancing Investment Analysis: Optimizing AI-Agent Collaboration in Financial Research}


\author{Xuewen Han}
\authornote{Both authors contributed equally to this research.}
\affiliation{%
  \institution{Tsinghua University}
  \city{Beijing}
  \country{China}
}
\email{hanxw19@mails.tsinghua.edu.cn}

\author{Neng Wang}
\authornotemark[1]
\affiliation{%
  \institution{AI4Finance Foundation}
  \city{Los Angeles}
  \state{California}
  \country{USA}}
\email{nengwang19@ucla.edu}

\author{Shangkun Che}
\affiliation{%
    \institution{Tsinghua University}
  \city{Beijing}
  \country{China}
}
\email{csk19@mails.tsinghua.edu.cn}

\author{Hongyang Yang}
\affiliation{%
 \institution{AI4Finance Foundation}
 \city{New York}
  \country{USA}}
 \email{hy2500@columbia.edu}

\author{Kunpeng Zhang}
\affiliation{%
  \institution{University of Maryland, College Park}
  \city{College Park}
  \state{Maryland}
  \country{USA}}
\email{kpzhang@umd.edu}

\author{Sean Xin Xu}
\affiliation{%
 \institution{Tsinghua University}
  \city{Beijing}
  \country{China}
}
\email{xuxin@sem.tsinghua.edu.cn}


\begin{abstract}


In recent years, the application of generative artificial intelligence (GenAI) in financial analysis and investment decision-making has gained significant attention. However, most existing approaches rely on single-agent systems, which fail to fully utilize the collaborative potential of multiple AI agents. In this paper, we propose a novel multi-agent collaboration system designed to enhance decision-making in financial investment research. The system incorporates agent groups with both configurable group sizes and collaboration structures to leverage the strengths of each agent group type. By utilizing a sub-optimal combination strategy, the system dynamically adapts to varying market conditions and investment scenarios, optimizing performance across different tasks. We focus on three sub-tasks: fundamentals, market sentiment, and risk analysis, by analyzing the 2023 SEC 10-K forms of 30 companies listed on the Dow Jones Index. Our findings reveal significant performance variations based on the configurations of AI agents for different tasks. The results demonstrate that our multi-agent collaboration system outperforms traditional single-agent models, offering improved accuracy, efficiency, and adaptability in complex financial environments. This study highlights the potential of multi-agent systems in transforming financial analysis and investment decision-making by integrating diverse analytical perspectives.

\end{abstract}



\received{3 Aug 2024}
\received[accepted]{28 Sep 2024}

\maketitle

\section{Introduction}
In recent years, generative artificial intelligence (GenAI) has made significant strides in various domains, including financial analysis and investment decision-making. The use of AI agents to analyze financial reports and aid in investment strategies has shown promising results, enhancing both accuracy and efficiency \cite{nie2024survey,kim2024financial,wu2023bloomberggpt,yang2023fingpt}.

Existing literature primarily focuses on single AI agents in financial analysis \cite{zhang2024finagent,liuai,yu2024finmem}. While these models have demonstrated effectiveness, there is growing interest in exploring the potential benefits of multi-agent systems. 
Current research has explored multi-agent debates (MAD) to improve agents' performance \cite{du2023improving} and shown improvements. 
However, applying MAD is not practical in larger agent groups. In this situation, agent collaboration can better leverage different agents' capabilities and improve task completion. This requires properly defined agent roles and a validated well-designed collaboration model. 

This research aims to fill this gap by investigating the effects of varying AI agent structures, with single to multiple agents, on financial analysis tasks. To address these questions, this paper focuses on the scenario of AI-powered investment research in the financial industry, which uses artificial intelligence to analyze financial market data, company financial reports, news, social media, and other sources, providing in-depth market insights and investment research reports. 
Specifically, we examine the 2023 SEC 10-K forms of 30 companies in the Dow Jones Index in terms of three primary tasks: fundamental analysis, market sentiment analysis, and risk analysis. These analyses inform final investment decisions, including whether to buy or not stocks and setting target prices for one week ahead. 
Primarily, we focus on two main configurations: 1) \textit{Agent Group Size}, to evaluate the impact of agent group size on financial annual report analysis; 2) \textit{Agent Collaboration Structure}, to identify the optimal collaboration structure for multi-agent groups for different financial tasks. 

Our study shows that for relatively simple tasks such as fundamental and market sentiment analysis, a single agent performs better than multiple agents from both investment research and Artificial Intelligence Generated Content (AIGC) perspectives. 
In contrast, multi-agents are more suitable for relatively complex tasks like risk analysis.
For collaboration, a structure where all agents communicate with each other performs better at simple tasks. Structure with absolute leadership optimizes efficiency and streamlines workflows, which is particularly suitable for scenarios requiring unified instructions and execution. The key contributions of this study can be summarized below: 

\begin{itemize}
    \item \textbf{Innovative Multi-Agent Collaboration Frameworks}: We defined several novel AI agent architectures from single-agent to multi-agent, and explored vertical/horizontal/hybrid multi-agent collaboration structures distinguished between internal implementations of leadership and cooperation.

    \item \textbf{Detailed Analysis of Information Flow}: We performed an in-depth analysis of the forms of information flow between agents in tasks of varying difficulty, explaining the core differences produced by different collaboration architectures. 
    
    \item \textbf{Optimal Agent Group for Investment Research}: Based on our experimental results, we identified the optimal multi-agent architectures for various sub-tasks. We demonstrated that an ensemble structure involving multiple agent groups achieves superior performance in investment research tasks, achieving target price prediction with 2.35\% avg. diff. and 66.7\% accuracy in buy or not decision-making, surpassing all other agent architectures.
\end{itemize}

\section{Related Works}

\subsection{LLM and AI Agents}
Large Language Models (LLMs) are advanced AI systems designed to process and generate human language \cite{zhao2023survey,radford2018improving,brown2020language}, leveraging deep learning architectures like transformers to capture complex linguistic patterns \cite{vaswani2017attention}. These models, exemplified by OpenAI's GPT-4 and Google's BERT  \cite{floridi2020gpt,devlin2018bert}, excel in diverse tasks such as translation, summarization, and conversational interfaces due to their ability to understand context and semantics. LLMs are trained on extensive datasets, which allows them to learn grammar, facts, and reasoning, although this also necessitates substantial computational resources. AI agents, on the other hand, are autonomous systems that interact with their environment to achieve specific objectives, often using LLMs to enhance natural language processing capabilities \cite{zeng2023large,durante2024agent,wang2024survey}. Applications range from personal assistants like Siri and Alexa to autonomous vehicles and healthcare diagnostics \cite{li2024personal,schmidgall2024agentclinic}. These agents significantly improve efficiency and productivity in various domains; for instance, they can automate routine tasks, allowing professionals to focus on more complex and creative work \cite{wang2024describe}. Furthermore, the financial sector is experiencing a profound transformation as AI agents are increasingly utilized in investment research \cite{nie2024survey}, enabling more sophisticated analysis and decision-making processes. 

\subsection{AI Agents in Financial Investment}
In the rapidly evolving field of financial investment research, Large Language Models (LLMs) and AI Agents are revolutionizing the way analysts and investors gather and interpret data \cite{harvel2024can,yang2023investlm,zhao2024revolutionizing,ko2024can}. These advanced technologies utilize natural language processing and machine learning to analyze vast amounts of unstructured data, such as news articles and financial reports, providing insights that enhance decision-making. Examples like StockAgent \cite{liuai}, FinAgent \cite{zhang2024finagent}, and FinMEM \cite{yu2024finmem} illustrate diverse applications of these technologies. StockAgent specializes in stock market analysis by predicting price movements, while FinAgent offers comprehensive financial assessments and strategic recommendations. FinMEM employs memory-enhanced modeling to integrate historical market knowledge with current conditions for long-term investment strategies. The benefits of using LLMs and AI Agents include increased efficiency, improved accuracy, and scalability across markets.

\begin{figure}[h]
  \centering
  \includegraphics[width=\linewidth]{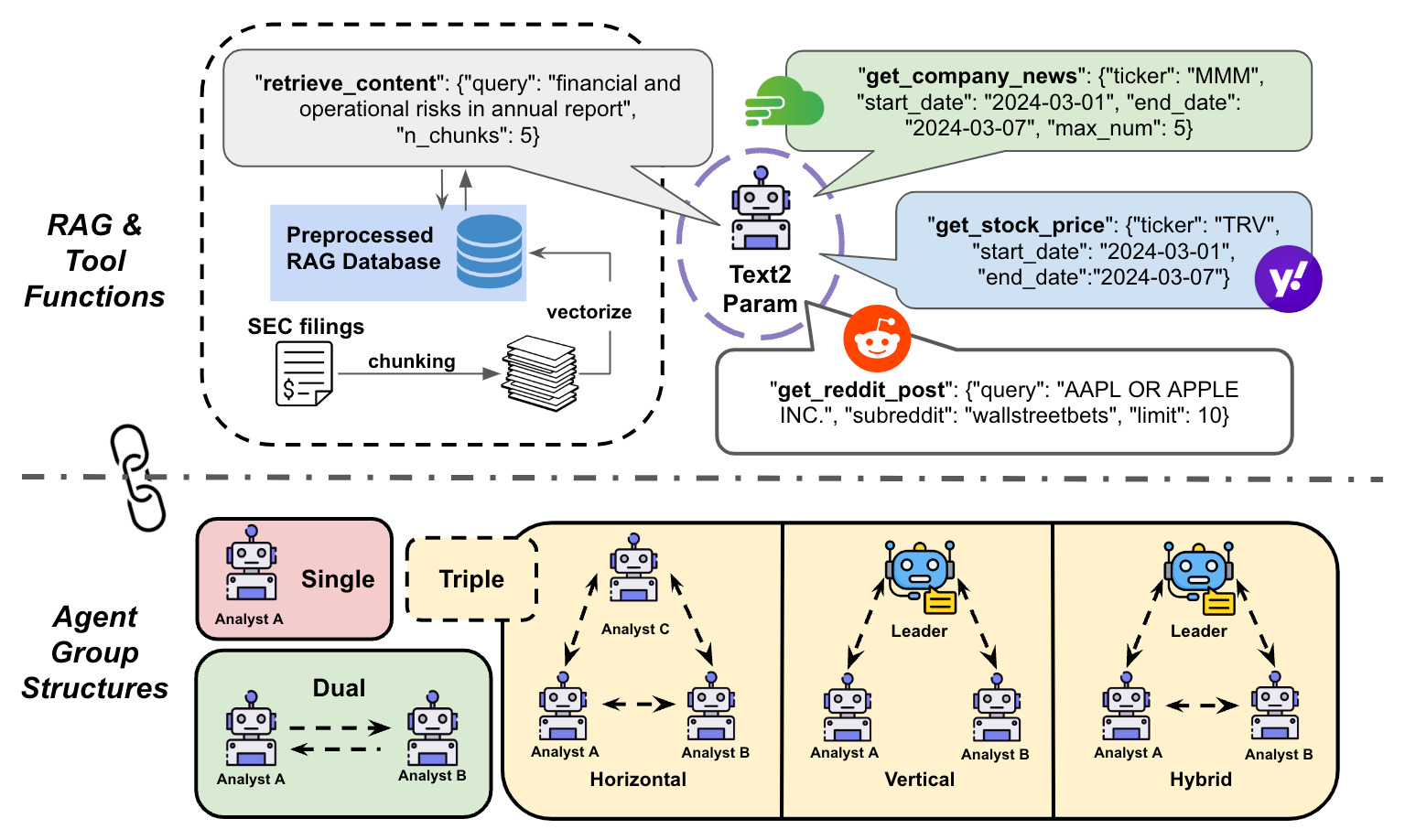}
  \caption{Overview of proposed multi-agent collaboration framework with unified RAG \& tool function calling.}
  \label{fig:overview}
\end{figure}

\section{Methodology}

\subsection{Overview} 
The architecture of AI agents varies from a single agent running independently to a collaborative form where multiple agents work together. Multi-agent systems involve two or more agents with the same or different LLM back-ends. In this section, we explore our methodology for building AI agent structures from single-agent to multi-agent, including multi-agent collaboration structures with the distinct internal implementation of leadership and cooperation. The overview of the structures is shown in Figure \ref{fig:overview}.

\subsection{Unified RAG \& Funtion Calling}
Relying on the Text2Param capabilities of the underlying LLM, it's not difficult for agents to invoke a variety of tools, such as retrieving stock prices from YFinance and fetching posts from Reddit's r/WallStreetBets. However, for investment research, analyzing company financial reports is also crucial. This requires equipping agents with the ability to retrieve relevant contexts from financial reports that are usually extremely long and complex. Typically, researchers adopt RAG (Retrieval-Augmented Generation) for such scenarios.

Unlike typical RAG use cases (e.g. QA), we do not manually provide query questions or adjust the retriever settings for optimal results. Instead, to fully evaluate the agent's capabilities, we have encapsulated the ability to retrieve chunks from the RAG database into a tool function as shown in Figure \ref{fig:overview}. Under this setup, agents can independently write the query and pick the number of retrievals based on task requirements. If the retrieved chunks do not meet the needs, the parameters could be refined and used for another round of recall. Such integration places the RAG functionality within a unified framework alongside other tool functions, making the scope of the agent's capabilities clearer and more manageable.

\subsection{Single-agent}

For a single-agent case study, we utilize a basic chatbot with the function-calling ability described in the previous section. System prompts define its role as an analyst and limit its behavior to write \& execute codes on its own. Through such constraints, it's easier to balance the difficulty level of each task based on not only the task itself but also the information available for each agent. There is no human intervention in any agent's process of handling the task, agents can decide to terminate their analysis on their own when they finish their tasks.

\subsection{Dual-agent Group}

In a group of two agents, agents' roles are similarly defined as, and have equal rights to conclude the given tasks. However, If we directly apply the implementation of a single-agent architecture, a common issue is that one agent might complete most of the work and then summarize and respond, skipping interaction with the other agent. Extra prompts are thus added to force communication between agents. For a dual group of Agents A\& B, the following responsibilities are added to Agent A's system prompt:

\begin{enumerate}
    \item Ask for advice from [Agent B] before you make any conclusion.
    \item Inspect analysis delivered by [Agent B] and give out advice.
    \item Reach a consensus with [Agent B] and provide the final analysis.
\end{enumerate}

\subsection{Triple-agent Group}

When the number of agents exceeds three, we can experiment with more complex and dynamic structures. These structures can be better adapted to scenarios involving larger groups or can function as unit structures within a larger organization. Centered around leadership and communication, we have developed three distinct agent collaboration patterns, which are referred to as vertical structure, horizontal structure and hybrid structure.

\subsubsection{Horizontal Collaboration}

In a horizontal multi-agent structure, all agents are on the same level. This architecture is easy to implement and expand and has been proven to be suitable for handling relatively simple tasks. In a horizontal agent structure, all agents equally participate and discuss tasks together. Their communication is shared, and each agent can see all messages from other agents. The horizontal agent structure is usually suitable for tasks requiring collaboration, feedback, and team discussion.

The implementation of a horizontal collaboration structure can be seen as an expansion of a dual-agent group, as similar prompts are utilized to force communication and information sharing. The dialogue unfolds in the form of a group chat, and the conversation in the group chat is visible to all roles. The speaking order follows a round-robin mechanism, with all roles taking turns to speak until any agent decides to end the conversation and report back to the user.

\subsubsection{Vertical Collaboration}

In the vertical multi-agent structure, the organization structure is leader-subordinate. Leaders are responsible for global planning and coordination, while subordinates handle specific data processing and analysis. Subordinates simply receive orders from the leader, finish the job assigned, and report back. They are not designed to communicate with other agents.

As such structure is not natively supported, we implemented it with a nested chat mechanism as shown in Algorithm \ref{algo:nested}. The leader needs to specify a subordinate's name at the end of the response and issue instructions to it. Such a format is captured by the framework, triggering a nested chat process towards the specified subordinate, starting with the leader's instruction. This nested chat content is not visible to other subordinates. After the nested chat ends, the subordinate's final response is delivered back to the leader. The leader then issues instructions continuously or integrates the subordinate's results to ultimately decide whether to end the entire conversation. To achieve this, a \textbf{leader\_prompt} is designed as below, where \textbf{group\_desc} introduces all subordinates and their responsibilities:

\newenvironment{custombox}[1]{
  \def\FrameCommand{
    \fboxsep=4pt 
    \fcolorbox{gray!80}{gray!10}}
  \MakeFramed{\FrameRestore}
  \noindent\textbf{#1} 
  \par\vskip6pt}{
  \endMakeFramed}

\smallskip
\begin{custombox}{Leader Prompt Sample}
    You are the leader of the following group: \textbf{\{group\_desc\}}
    
    As a group leader, you are responsible for coordinating the team's efforts to achieve the project's objectives. You must ensure that the team is working together effectively and efficiently. 
    \begin{itemize}[noitemsep, leftmargin=*]
        \item Summarize the status of the whole project progress each time you respond.
        \item End your response with an order to one of your team members to progress the project, if the objective has not been achieved yet.
        \item Orders should be follow the format: "[<name>] <order>".
        \item Orders need to be detailed, including necessary time period information, stock information, or instruction from higher-level leaders. 
        \item Make only one order at a time.
        \item After receiving feedback from a team member, check the results of the task, and make sure it has been well completed before proceeding to the next order.
    \end{itemize}
    
    Reply "TERMINATE" at the end when everything is done.
    
\end{custombox}

\subsubsection{Hybrid Collaboration}

In a hybrid multi-agent structure, agents have the same leveled roles and responsibilities defined as in the vertical structure but can communicate in a shared manner as in the horizontal structure. Unlike in the horizontal structure, where all agents have the right to report the final analysis and finish the conversation, lower-level agents here are asked to turn to the leader for the final analysis. This structure indicates a balance between leadership and equal cooperation.

Instead of using the leader\_prompt to establish leadership, two extra responsibilities are added separately to the leader and subordinates' system prompt as follows:

\begin{itemize}
    \item Leader
    \begin{enumerate}
        \item Give out tasks and advices to [Agent A] \& [Agent B].
        \item You should be the person to provide final analysis and finish the task.
    \end{enumerate}
    \item Subordinates
    \begin{enumerate}
        \item Report your findings to [Leader Agent] and ask for advices before providing final analysis.
        \item You're not allowed to finish the task without the permission from [Leader Agent].
    \end{enumerate}
\end{itemize}

The speaking order is still round-robin, but the conversation is consistently ended by the leader agent with the above prompts implemented.

\begin{algorithm}
\caption{Nested Chat Process}
\label{algo:nested}
\begin{algorithmic}[1]
    \REQUIRE User $U$, Leader Assistant $LA$, Initial Order $O$, Subordinates $\{S_1, S_2, \ldots, S_n\}$
    \ENSURE Final message after nested chats
    \STATE Initialize chat context $Context \gets [$O$]$
    \WHILE{True}
        \STATE $Response \gets$ Receive message from $LA$ based on $Context$
        \STATE Append $Response$ to $Context$
        \IF{$Response$ contains "TERMINATE"}
            \RETURN $Context$
        \ELSIF{$Response$ matches format \textless AgentName \textgreater: \textless Order \textgreater}
            \STATE $Agent \gets$ \textless AgentName \textgreater
            \STATE $Order \gets$ \textless Order \textgreater
            \IF{$Agent \in \{S_1, S_2, \ldots, S_n\}$}
                \STATE $NestedResponse \gets$ \textsc{NestedChat}($LA$, $Agent$, $Order$)
                \STATE Append $NestedResponse[-1]$ to $Context$
            \ENDIF
        \ENDIF
    \ENDWHILE
\end{algorithmic}
\end{algorithm}

\section{Experiments}

\subsection{Task Description}

This study focuses on the effectiveness of different AI agent structures in the context of investment research. Specifically, it examines the 2023 SEC 10-K forms of 30 companies in the Dow Jones Index to assess the performance of AI agents in three primary tasks: fundamentals analysis, market sentiment analysis, and risks analysis. These analyses inform final investment decisions, including whether to buy or sell stocks and setting target prices for one week ahead.

Each task is executed by a team of AI agents equipped with large language models. We evaluate three agent sizes: single, dual, and triple. Additionally, we examine three collaboration structures under the triple-agent setup: horizontal, vertical, and hybrid.

\subsection{Implementation Details}

The data set used for investment research includes the 2023 annual reports of the 30 Dow Jones Index companies, along with stock prices at the time of report release and one week later. As for information available through tool function calls, all the search ranges of either stock movements or social media are limited to before the release date of the report to prevent data leakage.

Experiments are conducted on the FinRobot platform developed by the AI4Finance community \cite{yang2024finrobotopensourceaiagent}. The communication is achieved with Microsoft's AutoGen's GroupChat \cite{wu2023autogen} for a basic horizontal group. For complex agent group structures that are not natively supported by AutoGen, like vertical \& hybrid collaboration, we add leadership with our own implementation. Codes are available on our GitHub Repository\footnote{\href{https://github.com/AI4Finance-Foundation/FinRobot}{https://github.com/AI4Finance-Foundation/FinRobot}}.

This study uses the "GPT-4-1106-vision-preview" model provided by OpenAI as the foundation for all AI agents. Tool functions are also transformed into JSON schema as defined by the OpenAI API.

All agents are equipped with Retrieval-Augmented Generation (RAG) capabilities as a tool function for recalling context from SEC filings based on keywords or key phrases. Texts from the annual reports are parsed from PDF files, divided into chunks of 1,000 tokens and then vectorized using the "all-MiniLM-L6-v2" model of sentence transformers \cite{reimers-2019-sentence-bert}, which is also utilized for queries during retrieval.

Different sets of additional tool functions are provided for different tasks. For fundamental analysis, we leverage several APIs of Financial Modeling Prep (FMP) \cite{Vulcain2023FinancialModelingPrepAPI} for company fundamentals and the YFinance package \cite{ValueRaider2024ranaroussi} for more stock-related information. For market sentiment analysis, we provide agents with encapsulated FinnHub API \cite{Vu2024Finnhub} for company-related market news, and Python Reddit API Wrapper (PRAW) \cite{Boe2024praw} to gather relevant social media posts. For risk analysis, we provide no additional tool function thus making the task harder.

\subsection{Evaluation}

To evaluate the performance of AI agents in tasks, we need to analyze and assess AI-generated content.

From the functional perspective, the AI-generated content in this study is used as reports to provide intelligent investment research services, which should meet the requirements of research reports in the financial investment field, such as information accuracy, logical completeness, professional evaluation indicators, and the ability to guide investment decisions.

From the perspective of Artificial Intelligence Generated Content (AIGC), which is another category widely studied in academia, mainly focusing on evaluating and analyzing content generated by AI, we need to pay attention to the readability, coherence, and other general indicators of the generated text, adopting these indicators to assess whether the AI-generated research reports regarding text quality are close to the human level.

In summary, from the perspective of investment research reports as well as AIGC, totally we have 7 indicators, summarized in Table ~\ref{tab:indicators}. 

\subsubsection{Evaluation Indicators from the Perspective of Investment Research Reports}

From a functional perspective, we focus on the quality of AI-generated content as a research report. Since we divided investment analysis into three sub-tasks and combined the three sub-modules to complete the final research report and investment suggestions, we designed evaluation indicators for both the overall research report and the three sub-tasks:

\begin{enumerate}
    \item For fundamental analysis sub-task, calculating standard indicators is a common practice in the financial industry \cite{abarbanell1997fundamental}. Therefore, we need to assess whether the AI agent correctly leverages indicators to provide a comprehensive analysis of the company's fundamentals. 
    \item For market sentiment analysis sub-task, research shows that market sentiment significantly affects investor behavior and investment decisions \cite{bollen2011twitter}. Therefore, we assess the quality of AI-generated content by analyzing market behavior and judging market trends based on market sentiment. 
    \item For risk analysis sub-task, considering the effectiveness of risk analysis in investment performance \cite{hull2014evaluation}, we assess whether this part of the AI-generated content includes sufficient potential risk identification and provides effective suggestions.
    \item For the evaluation of overall research reports and investment suggestions, since the prediction of investment results directly affects returns\cite{ertimur2007measure}, we evaluate the accuracy of predicted one-week target stock price and the buy/not buy investment suggestions based on this prediction.
\end{enumerate}

\subsubsection{Evaluation Indicators from the Perspective of AIGC}

From the perspective of AIGC, related research has proposed evaluation indicators such as relevance, attractiveness, coherence, and readability, where 
readability and coherence are suitable for the scenarios in this study \cite{zhang2023artificial,zeng2024large}. 
Readability refers to the precision and natural expression of the content. Coherence refers to whether the answer's content is organized, grammatically correct, and easy to understand. Therefore, in this study, we will evaluate the analysis reports generated in sub-tasks from these two aspects.

\begin{table}[ht]
\centering
\caption{Evaluation Indicators}
\label{tab:indicators}
\begin{tabularx}{0.45\textwidth}{@{}l|X@{}}
\toprule
\multicolumn{2}{c}{\textbf{The Perspective of Investment Research Reports}} \\ \midrule
\multicolumn{2}{c}{(1) Indicators for Sub-tasks} \\ \midrule
\textbf{Fundamental} & Whether the AI-generated content considers commonly used model evaluation indicators and provides a comprehensive analysis. \\ \addlinespace
\textbf{Sentiment} & The quality of AI-generated content in analyzing market behavior and judging market trends based on market sentiment. \\ \addlinespace
\textbf{Risk} & Whether the AI-generated content includes sufficient potential risk identification and provides effective suggestions. \\ \midrule
\multicolumn{2}{c}{(2) Indicators for Overall Report} \\ \midrule
\textbf{Stock Price} & The accuracy of the one-week target stock price predicted by the AI. \\ \addlinespace
\textbf{Buy / Not Buy} & Accuracy of the buy/not buy investment suggestions based on the prediction. \\ \midrule
\multicolumn{2}{c}{\textbf{The Perspective of AIGC}} \\ \midrule
\textbf{Readability} & Precision and natural expression. \\ \addlinespace
\textbf{Coherence} & Whether the content is organized, grammatically correct, and easy to understand. \\ \bottomrule
\end{tabularx}
\end{table}

\subsection{Analysis of Experimental Results}

To evaluate AI-generated content, LLMs like GPT-4 are frequently utilized in research \cite{zeng2024large}. In this study, we adopt a similar approach by designing specific prompts to enable the GPT-4 model to evaluate text quality based on our defined criteria. The GPT-4 model is tasked with providing a score ranging from 1 to 5, with 1 indicating minimal task completion and 5 indicating full task completion. For example, for the "fundamental analysis" indicator, the prompt input to GPT-4 is: ``Please evaluate from the perspective of whether the fundamental indicators of the enterprise have been identified or discussed, eradicated and judgments or suggestions have been given. Please give a score of 1-5 based on your judgment. 1 represents a lack of fundamental analysis or nonsense, 3 represents a certain fundamental analysis capability but not sufficient, and 5 represents a relatively complete fundamental analysis.''

\subsubsection{Analysis of Sub-task Quality}

In this study, we analyze the annual reports of listed companies through three primary tasks: company fundamentals analysis, market sentiment analysis, and risk analysis. Each task follows a unified introduction and scoring system to ensure consistency.

\textbf{\textit{Company Fundamentals Analysis}} focuses on evaluating a company's financial health, operational efficiency, profitability, and cash flow status. This task involves a detailed examination of financial data and business operations from the annual report.

\textbf{\textit{Market Sentiment Analysis}} gathers information from diverse sources such as news, social media, and investor reports to assess sentiment changes, which are closely linked to market trends and investment opportunities. By detecting market sentiment, investors can swiftly identify market signals and anticipate potential risks and opportunities. 

\textbf{\textit{Risk Analysis}} involves identifying investment risk factors based on financial data. Identifying risk events allows investors to adapt their strategies flexibly, mitigating potential losses.

\begin{table}[h]
  \caption{Results of sub-tasks quality analysis (Size)}
  \label{tab:res_final}
  \begin{tabular}{cccc}
    \toprule
    Collaboration & Fundamental & Sentiment & Risk\\
    \midrule
    Single & \textbf{4.70} & \textbf{3.93} &3.57 \\
    Dual & 4.17& 3.90 & 3.77 \\
    Triple & 3.97 & 3.77 & \textbf{3.83} \\
  \bottomrule
\end{tabular}
\end{table}

\vspace{-0.5cm}
\begin{figure}[h]
  \centering
  \includegraphics[width=\linewidth]{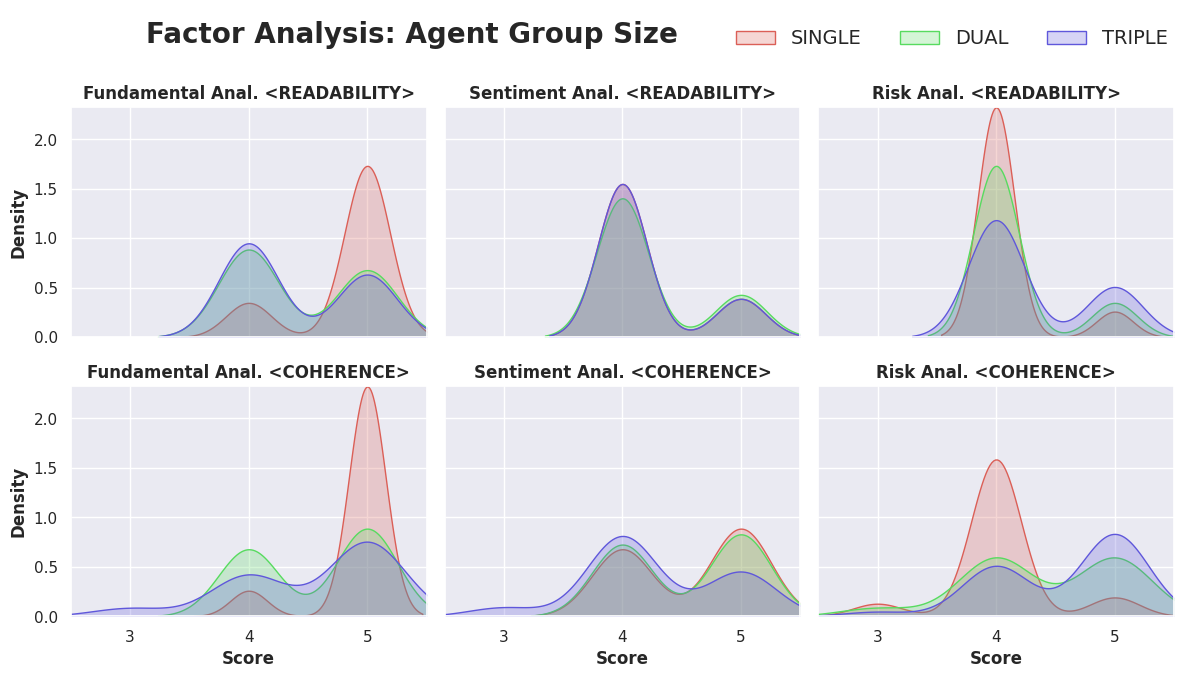}
  \caption{Evaluation of AI-generated content - by Group Size}
  \label{fig:agent_group_size}
\end{figure}

\subsubsection{Performance of Different Agent Group Sizes}

First, we compared the quality of generated content from the perspective of investment research reports. As illustrated in Table~\ref{tab:res_final}, for fundamental analysis, a single agent outperforms larger groups. As the size of the agent group increases, performance decreases. A similar trend is observed in the sentiment analysis task, though the performance gap between a single agent and larger groups is smaller. However, for risk analysis, a group of three agents performs the best, while a single agent shows the poorest performance. 

Next, we evaluated the quality from the perspective of AI generated content. The readability and coherence score distributions among different group sizes are illustrated in Figure \ref{fig:agent_group_size}. 
For fundamental and sentiment analysis tasks, a single agent excels in readability and coherence. However, for risk analysis, a group of three agents outperforms, while a single agent performs the worst in both readability and coherence metrics.
Findings in this part are consistent with the above evaluation from the perspective of investment research reports.

The complexity and depth of the tasks increase from fundamental analysis to sentiment analysis and risk analysis. In fundamental analysis, standard metrics such as financial health, operational efficiency, profitability, and cash flow status are used. Single agents, equipped with the necessary tools to calculate these metrics, can complete the task efficiently and produce high-quality reports. In contrast, larger agent groups introduce noise through debates and differing opinions, reducing the readability and coherence of the final reports.

Sentiment analysis based on social media presents a relatively complex task. Multiple sentiment tendencies can coexist in the market, and the content, particularly from individual forums, may be contradictory. Agents must analyze substantial information to provide a comprehensive sentiment analysis. Additionally, market sentiment is dynamic, requiring consideration of the time dimension for accurate conclusions. Despite these complexities, the task remains manageable due to restricted information sources, enabling a small number of agents to produce highly readable reports. Consequently, the performance of a single agent is not as distinctly superior as in fundamental analysis.

Risk analysis is highly challenging, requiring agents to identify potential risks in annual reports, online news, and social media. Collaborative efforts and debates among agents enhance insights and accuracy, resulting in more comprehensive risk reports. Larger agent groups can leverage diverse perspectives and expertise, leading to more thorough identification and analysis of risks. The collaborative process improves the overall quality of the report, making it more readable and coherent. Therefore, larger agent groups perform better in this task, highlighting the importance of collaboration in handling complex and multifaceted analyses. Additionally, the success of larger agent groups in risk analysis emphasizes the value of diversity in thought and expertise when dealing with intricate and dynamic information. The ability to pool knowledge and cross-verify findings helps mitigate individual biases and leads to more robust and reliable conclusions. This collaborative approach is particularly beneficial in scenarios where the stakes are high, and the accuracy of the analysis is crucial for informed decision-making.

\begin{table}[h]
  \caption{Results of sub-tasks quality analysis (Multiple)}
  \label{tab:res_final2}
  \begin{tabular}{cccc}
    \toprule
    Collaboration & Fundamental & Sentiment & Risk\\
    \midrule
    Vertical & 3.20& 3.43 & \textbf{4.23} \\
    Horizontal & 3.97 & \textbf{3.77} &3.83 \\
    Hybrid & \textbf{4.03} & \textbf{3.77} &3.72 \\
  \bottomrule
\end{tabular}
\end{table}

\vspace{-0.5cm}
\begin{figure}[h]
  \centering
  \includegraphics[width=\linewidth]{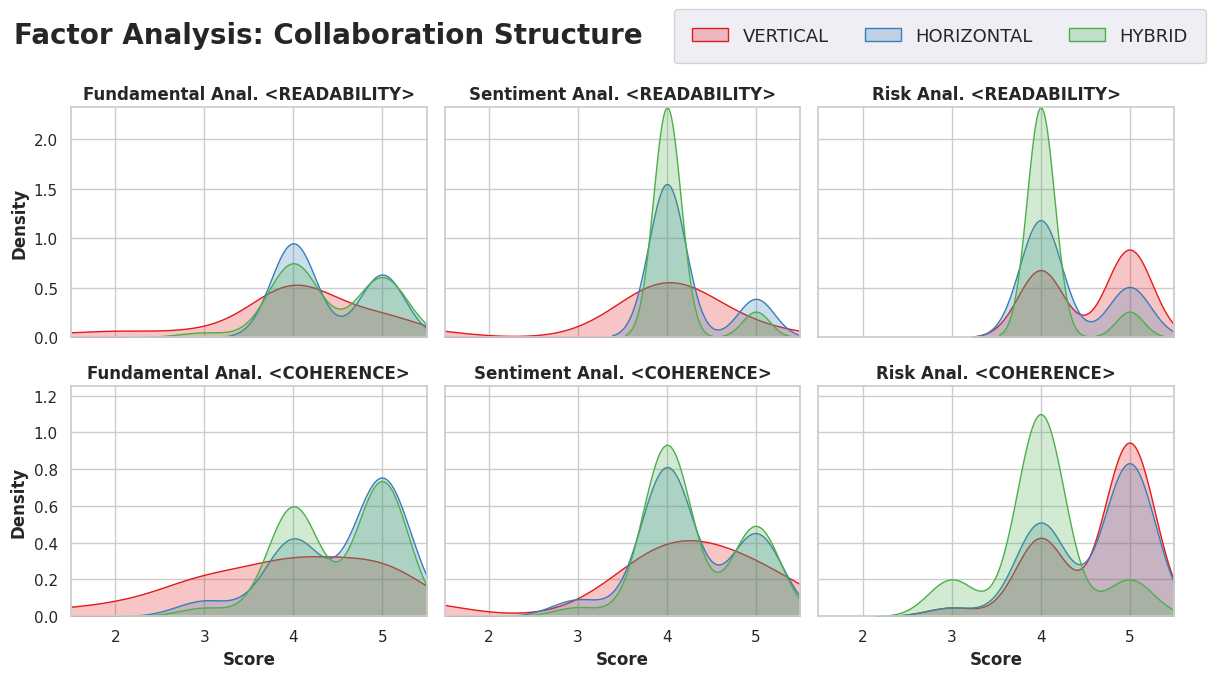}
  \caption{Evaluation of AI-generated content - by Collaboration Structure}
  \label{fig:collaboration_structure}
\end{figure}

\subsubsection{Performance of Different Agent Collaboration Structures}

\begin{figure}[h]
  \centering
  \includegraphics[width=\linewidth]{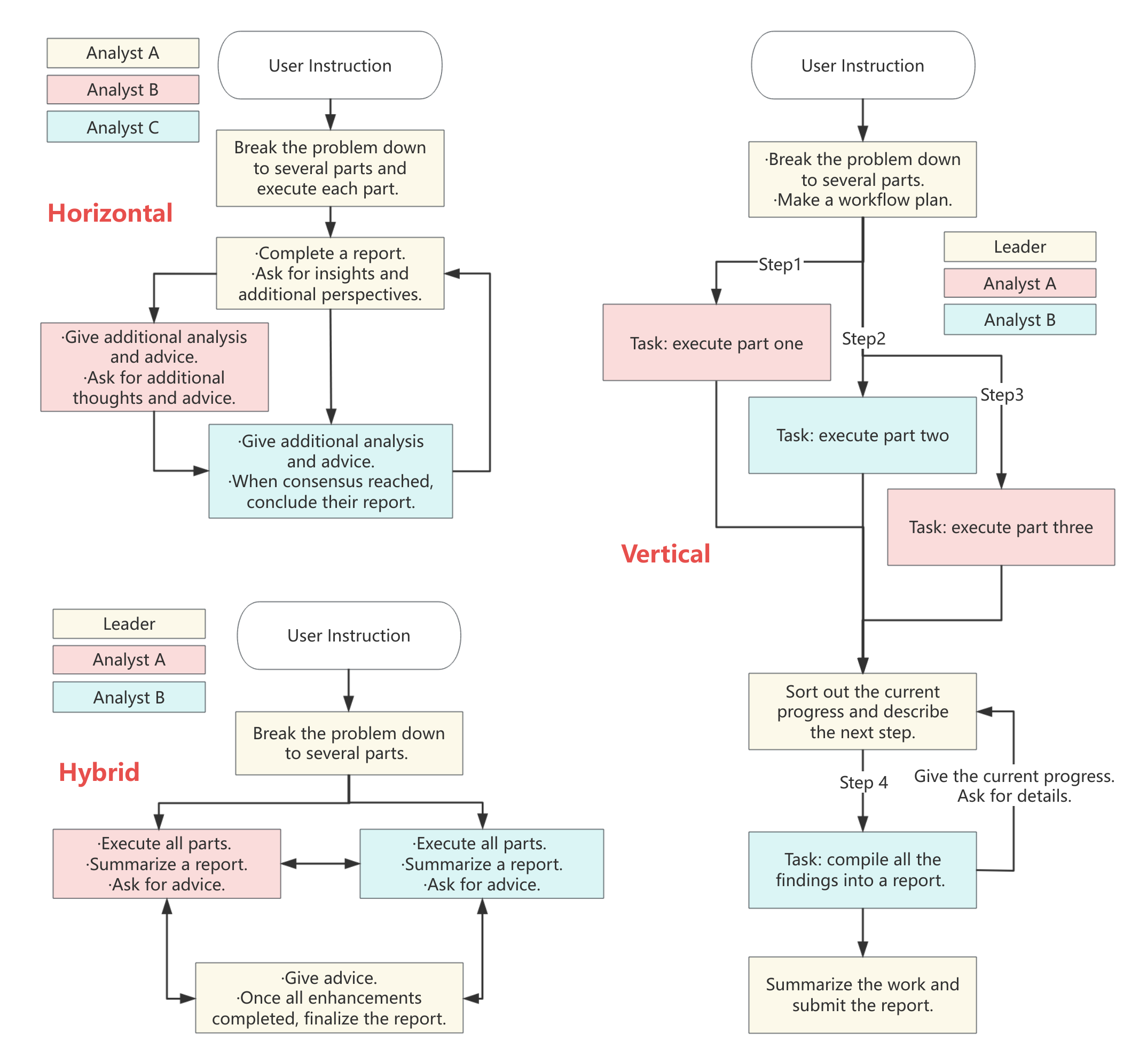}
  \caption{The information flow between agents.}
  \label{fig:last}
\end{figure}

Furthermore, we conduct the same generated content quality evaluation based on different agent collaboration structures. The overall trend in Table~\ref{tab:res_final2} indicates that for easier tasks, such as fundamental and sentiment analysis, hybrid and horizontal structures perform better. For more complex tasks, like risk analysis, the vertical structure yields superior performance. We carefully analyzed the execution and communication records of agents of all structures in various tasks, and drew the information flow between agents in these structures shown in Figure~\ref{fig:last}.

Among the three sub-tasks, both fundamental analysis and sentiment analysis have clear and relatively simple task objectives and execution processes. For example, fundamental analysis mainly involves extracting key fundamental analysis indicators from annual reports, which are common indicators used in financial investment. Sentiment analysis first crawls data from multiple media data sources, and then further summarizes the sentiment in the text. It can be summarized that these two tasks mainly involve the collection, organization and summary of key information, and ensure the comprehensiveness and reliability of information. For such relatively basic tasks, when multiple agents can communicate, multiple agents can share their own information, and the information provide by these agents can be used to correct errors and complement each other to improve the comprehensiveness and reliability.
The final decision-making power of the leader is not very useful for such information-collecting-and-summarizing and non-decision-making tasks. Therefore, we can observe that the horizontal structure and the hybrid structure perform better in these two tasks because they allow communication and collaboration between agents.

As shown in the information flow in Figure~\ref{fig:last}, there is communication between agents in the horizontal structure and the hybrid structure. One agent can make supplementary analysis or give suggestions on the results of other agents. When the agents reach a consensus, they will output the final report. For example, in the market sentiment analysis task for Honeywell, we found that in the horizontal and the hybrid structure, when analyst A completed the analysis task, it would ask for suggestions from other agents: ``I would appreciate your insights and any additional perspectives you might have before finalizing our report''. Then analyst B would give effective suggestions: ``It's also worth considering the broader industry trends and economic factors that may influence investor sentiment towards Honeywell. For instance,...''. Through collaboration, the completion of such information collection and organization tasks will be more comprehensive and reliable.

Unlike these two tasks, risk analysis is more inclined to an analysis task and a decision-making task. It not only requires the above-mentioned collection of various types of information, but more importantly, it requires in-depth analysis of various information and judges potential investment risks based on the existing knowledge of the agent, which requires the agent to have analysis and judgment capabilities. Risk evaluation is often implicit, and different agents may have differing opinions. For this task, we need a core leader to synthesize the analysis results of other analysts and make the final risk judgment. It is very important for different analysts to provide valuable insights from their own perspectives. 
If analysts can communicate with each other, it may lead to convergence of opinions among analysts, making the information available to decision makers biased. 
The vertical structure, through centralized control, ensures efficiency and consistency in information processing. The leading agent evaluates and filters information collected from subordinate agents, focusing on the most critical data. The leader’s role in synthesizing diverse viewpoints helps maintain coherence and focus in the final report. The horizontal structure, where all agents are on the same level and freely communicate, may lead to unresolved disagreements and introduce noise into the final report. In the hybrid structure, although there is a leader, the communication between the two agents can complicate the leader's decision-making process, potentially affecting the quality of the final report.
Therefore, for this kind of task, we found that the vertical structure performs better.

As shown in the information flow in Figure~\ref{fig:last}, in a vertical structure, the leader can divide the task into multiple steps, assign each step to different analysts, and make the final judgment. Analysts are not allowed to communicate with each other, so they can freely analyze potential risks from different angles. For example, in the risk analysis of IBM, analyst A qualitatively concluded based on the analysis of revenue trends, profit margins, debt levels, operational challenges, and market risks: ``IBM faces competition and must navigate a rapidly evolving technology landscape to sustain growth and productivity. Given these considerations, I would assign IBM a risk score of 7 out of 10''.
Analyst B use various information to establish a risk assessment model, combined with quantitative financial data and various factors, and set weights for different factors based on the importance of each factor. He calculated a risk score of 6.35/10 through quantitative analysis methods: ``The model will use a weighted scoring system to quantify the impact of each risk factor on IBM's overall risk profile. Each risk factor will be assigned a weight based on its perceived impact on IBM's operational and financial health''.
Finally, the leader combines the qualitative information given by Analyst A with Analyst B's risk assessment model. The rationality of factor selection and weight setting in B's quantitative analysis model is analyzed, and the final risk judgment is given: ``It is recommended to decrease the risk score from 7 to approximately 6.35 to more accurately reflect IBM's risk profile''.

Further discussion highlights that the choice of collaboration structure significantly impacts the quality of generated content. For fundamental and sentiment analysis, hybrid and horizontal structures are beneficial due to their support for open dialogue and diverse perspectives. For risk analysis, a vertical structure with a decisive leader is advantageous. The leader’s ability to integrate different viewpoints and make final decisions facilitates supervision and control, ensuring the coherence and readability of the report. This underscores the importance of selecting an appropriate collaboration structure based on the nature and complexity of the task to optimize the quality of the generated content.



\subsubsection{Financial Decision Making Analysis}
Lastly, we perform a financial decision analysis by integrating three key sub-tasks. This integration aims to provide a comprehensive evaluation of our system’s performance. We leverage GPT-4 to predict stock prices for the upcoming week with the information provided by three sub-tasks and provide investment recommendations. On the one hand, we utilize the same structure-generated content to generate final decision. On the other hand, our sub-task analysis reveal that vertical agent structure excels in risk analysis, while single agent structure performs better in fundamental and market sentiment tasks. Consequently, we select the optimal structure for each sub-task, resulting in a “ensemble structure.”

First, we evaluate the error between the target price predicted by GPT-4 for the upcoming week and the actual price. The error is defined as the absolute difference between the predicted and actual price, divided by the actual price. Despite the smallest target price difference is observed in dual agent structures, the ensemble agent structure outperforms other structures in price prediction accuracy.

Then, we focus on the core evaluation metric of for intelligent investment research: the accuracy of AI agents in making investment decisions, specifically determining whether to buy or not based on predicted stock price movements. In our experiment involving 30 stocks, the ensemble structure successfully predicted the movements of 20 stocks, achieving a prediction accuracy of 66.7\%, whereas the full vertical agent structure only achieved 50\%. This demonstrates that different agent structure designs significantly impact the performance of intelligent investment research. Our findings suggest that agents with ensemble structures are more suitable for this practical application. This also highlights the importance of designing multi-agent structures tailored to the specific sub-tasks involved in financial decision-making.

\begin{table}[h]
  \caption{Results of CIO's final investment recommendation}
  \label{tab:res_final3}
  \begin{tabular}{ccl}
    \toprule
    Collaboration & Avg. Diff. to Target & Binary Acc.\\
    \midrule
    Single & 2.43\% & 63.3\% \\
    Dual & \textbf{2.24}\% & 63.3\% \\
    Vertical & 4.75\% & 50.0\% \\
    Horizontal & 2.50\% & 63.3\% \\
    Hybrid & 2.57\% & 56.7\% \\
    Ensemble & \textbf{2.35\%} & \textbf{66.7\%} \\
  \bottomrule
\end{tabular}
\end{table}

\section{Conclusion}

This paper investigates the impact of AI agent structures on financial document analysis tasks. We designed AI agents of varying sizes: single, dual, and triple. Additionally, we implemented different collaborative structures among multiple AI agents, namely horizontal, vertical, and hybrid configurations. Our experiments, conducted on the 10-K forms of 30 companies listed on the Dow Jones Index, reveal the heterogeneous performance of different AI agent structures across various tasks. The results highlight the importance of suitable AI agent structure selection in emphasizing financial documents. Overall, our findings provide valuable insights into the optimization of AI agent structures for financial document analysis, offering practical guidelines for enhancing performance through strategic design and collaboration.

This work is one step toward financial document analysis for financial tasks leveraging AI agents. Future research in this area may include: ensuring the continuous and complete flow of information in an extensively hierarchical multi-agent system; leveraging small-scale debates in group structure to enhance task performance; replacing the sub-optimal structure selection/combination strategy with a globally optimized task approach.

\bibliographystyle{ACM-Reference-Format}
\bibliography{ref}

\end{document}